\documentclass[10pt,twocolumn,letterpaper]{article}

\usepackage{iccv}              

\usepackage{multirow}
\usepackage{array}
\usepackage{pifont}
\usepackage{tabularx}
\usepackage{makecell}
\usepackage{amsmath}
\usepackage{indentfirst}
\usepackage{siunitx}
\usepackage{algpseudocode}
\usepackage[linesnumbered,ruled,vlined]{algorithm2e}
\usepackage{csquotes}
\usepackage{xcolor}

\definecolor{iccvblue}{rgb}{0.21,0.49,0.74}
\usepackage[pagebackref,breaklinks,colorlinks,allcolors=iccvblue]{hyperref}

\def\paperID{7303} 
\def\confName{ICCV}
\def\confYear{2025}

\title{Constraint-Aware Feature Learning for Parametric Point Cloud}

\author{Xi Cheng$^{1}$ \quad
Ruiqi Lei$^{1}$ \quad
Di Huang$^{1}$ \quad
Zhichao Liao$^{1}$ \quad
Fengyuan Piao$^{1}$ \quad
Yan Chen$^{1}$ \\
Pingfa Feng$^{1, 2}$ \quad
Long Zeng$^{1, \dag}$ \\
$^{1}$Tsinghua Shenzhen International Graduate School, Tsinghua University, Shenzhen, China\\
$^{2}$Department of Mechanical Engineering, Tsinghua University, Beijing, China\\
\tt\small \{chengxi23, leirq22\}@mails.tsinghua.edu.cn, dihuangdylan@gmail.com,\\ 
\tt\small \{liaozc23, pfy22, chenyan23\}@mails.tsinghua.edu.cn,\\
\tt\small fengpf@tsinghua.edu.cn, zenglong@sz.tsinghua.edu.cn
}

\begin{document}
\maketitle

\renewcommand{\thefootnote}{}
\footnotetext{${\dag}$: Corresponding author.}
\renewcommand{\thefootnote}{\arabic{footnote}}  

\begin{abstract}

\noindent Parametric point clouds are sampled from CAD shapes and are becoming increasingly common in industrial manufacturing. Most CAD-specific deep learning methods focus on geometric features, while overlooking constraints inherent in CAD shapes. This limits their ability to discern CAD shapes with similar appearances but different constraints. To tackle this challenge, we first analyze the constraint importance via simple validation experiments. Then, we introduce a deep learning-friendly constraints representation with three components, and design a constraint-aware feature learning network (CstNet), which includes two stages. Stage 1 extracts constraint representation from BRep data or point cloud based on local features. It enables better generalization ability to unseen dataset after pre-training. Stage 2 employs attention layers to adaptively adjust the weights of three constraints' components. It facilitates the effective utilization of constraints. In addition, we built the first multi-modal parametric-purpose dataset, i.e. Param20K, comprising about 20K CAD instances of 75 classes. On this dataset, CstNet achieved 3.49\% (classification) and 26.17\% (rotation robustness) accuracy improvements over the state-of-the-art. To the best of our knowledge, CstNet is the first constraint-aware deep learning method tailored for parametric point cloud analysis. Our project page with source code is available at: \url{https://cstnetwork.github.io/}.

\end{abstract}
    
\section{Introduction}
\label{sec:intro}

\begin{figure}[ht]
  \centering
  \includegraphics[width=\linewidth]{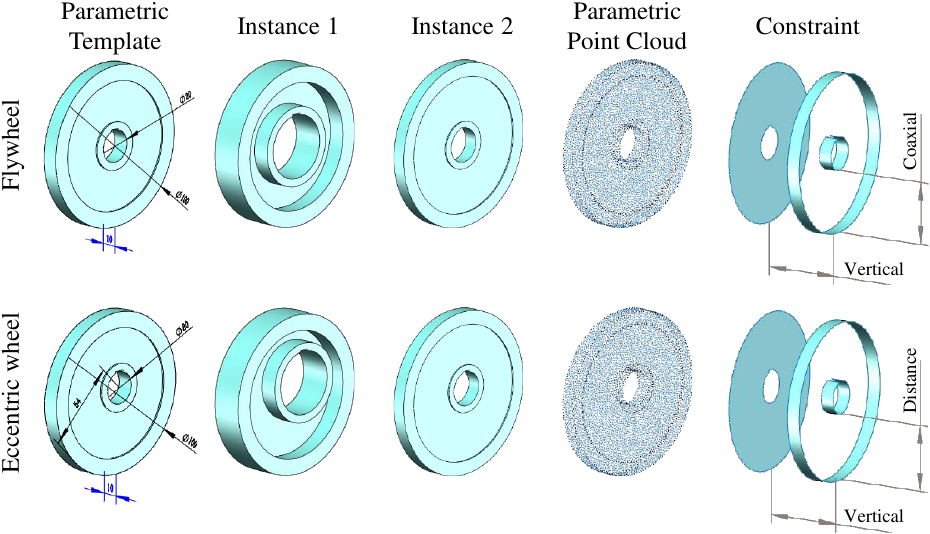}
  \caption{\textbf{Motivation.} Some CAD shapes are visually similar but serve different functions. The eccentric wheel has a distance constraint, \ie its rotation axis is offset from outer cylindrical surface. It is used for precise positioning or tension adjustment. In contrast, the flywheel has a coaxial constraint, \ie its rotation axis is coaxial with outer cylindrical surface. It is used to store rotational kinetic energy. While these shapes are challenging to distinguish visually, they can be effectively differentiated from constraints.}
  \label{fig:our_inov}
\end{figure}

\noindent Parametric point clouds are sampled from CAD shapes or captured from various industrial scenarios by common digital devices. As illustrated in \cref{fig:our_inov}, CAD shapes are typically instantiated from parametric templates, which comprise primitives, constraints, and shape parameters. By specifying a series of parameter configurations, a template can be instantiated into a family of CAD shapes \cite{zl1}. In industrial domain, many CAD shapes are visual similar but have different constraints in their functional regions, examples shown in \cref{fig:our_inov}. This poses challenges for geometric features-based point cloud deep learning methods. Thus, a constraint-aware feature learning method for parametric point cloud is fundamental for a wide range of industrial applications, such as engineering model design \cite{zl1, ref36}.

First, most existing studies \cite{hpnet, parsenet, spfn, h3d} focus on primitive regularity (\eg uniform appearance and distinct boundaries). They usually leverage geometric features only, but ignore constraints between primitives. Although these methods have been effectively applied in point cloud segmentation and primitive reconstruction, they inherently limit their discrimination capability for CAD shapes that are visually similar yet functionally distinct.
While some studies discussed constraints in 2D engineering sketches \cite{cad_lang, ref47}, the considered primitives in 2D sketches are supposed available and generally few in number. Thus, these approaches can be hardly extended to 3D parametric point clouds. 
Second, most CAD-specific methods rely on specialized datasets with additional labels (\eg primitive, parameter, and normal labels). However, point clouds are typically acquired via LiDAR or RGB-D cameras, and in most cases only contain coordinate information. It thus restricts the practical applicability of these approaches. 
Third, there are still numerous challenges in constraint-aware point cloud analysis. Constraints represent the spatial relationships between primitives, but the number and types of primitives cannot be directly obtained from point clouds. Predicting primitives from point clouds also poses challenges, the point cloud unordered nature results in unordered outcomes, leading to a mismatch between the predictions and labels. In addition, processing relations between every primitive pairs would produce an overwhelming data volume, since some CAD shapes contain many primitives. Assuming that the constraints have been obtained, effectively leveraging them remains a significant challenge. 
 
To address the aforementioned issues, we transform the traditional constraint representation (\eg constraints in SolidWorks or UG \cite{gcst}) into three components: primitive's Main Axis Direction (MAD), Adjacency (Adj), and Primitive Type (PT). Representing constraints with MAD-Adj-PT converts the relations between primitive pairs into relations between primitives and a common reference, thereby significantly reducing the data volume. 
Moreover, MAD-Adj-PT can be inferred by local point patches, and the local features of CAD shapes are predominantly composed of primitives such as planes and cylinders. Consequently, a model that relies solely on local features can generalize to unseen datasets after pre-trained, this property eliminates the reliance on constraint labels during practical applications. 
To mitigate the impact of point cloud unordered nature, we encode MAD-Adj-PT as point-wise features. As a result, each point with constraint representation in point cloud is defined as $(x, y, z, MAD, Adj, PT)$.
The aforementioned processing transform the traditional constraint representation into deep learning-friendly form.
After that, we designed the Constraint-Aware Feature Learning Network (CstNet), which consist of two stages. Stage 1 is designed to obtain MAD-Adj-PT. The architecture design is based solely on point cloud local features, experiments show that it exhibits strong generalization ability on unseen datasets, and its training data can be easily obtained by processing unlabeled BRep data. Following MAD-Adj-PT acquisition, Stage 2 separately extract features from MAD-Adj-PT, and apply attention layers to weight these features, thereby facilitating their effective utilization.

We quantitatively evaluate CstNet on ABC \cite{ref51} and Param20K dataset: MAD-Adj-PT prediction, classification, and rotation robustness evaluation. For MAD-Adj-PT prediction, CstNet surpasses the state-of-the-art (SOTA) by around 5\% and 14\% instance accuracy in ABC (test set) and Param20K (unseen dataset), respectively. For classification and rotation robustness experiments on Param20K, CstNet outperforms the SOTA by 3.49\% and 26.17\% instance accuracy, respectively. 

In summary, our contributions are threefold:

\begin{itemize}
    \item We introduced a deep learning-friendly constraint representation and designed the first constraint-aware feature learning network CstNet for effective constraint acquisition and leveraging for parametric point cloud. 
    
    \item We built the first parametric-purpose Param20K dataset, which consists of about 20K CAD instances of 75 classes. Each instance include mesh, point cloud, and BRep data. Given the current scarcity of labeled CAD shape datasets \cite{ref53, ref54}, Param20K offers a solid basis for future studies in parametric CAD domains.

    \item We conducted extensive experiments on constraint acquisition, classification, robustness evaluation and ablation studies. The performance of CstNet has a significant improvement compared with SOTA which usually focus on geometric features only. 

\end{itemize}

\section{Related Work}
\label{sec:formatting}

\noindent{\textbf{General point cloud analysis. }}View-based methods typically project the point cloud from multiple views, the obtained figures are then fed in convolutional neural networks (CNNs) for feature extraction \cite{ref4, ref5, ref6, ref7, ref8, ref9}. How to address information loss during the projection and optimizing viewpoints plays a critical role.
Voxelization-based methods convert point clouds into 3D voxels, and then employ 3D CNNs for feature extraction \cite{ref10, ref11, ref12, ref13, ref14, ref15}. The size and orientation of 3D voxels significantly influences the accuracy of these methods, and information loss occurs during the conversion to voxels.
Point-based methods directly use points as input, thereby avoiding information loss associated with view-based and voxel-based methods, it is also the most popular point cloud learning method \cite{ref1, ref2, ref3, ref16, ref17, ref18, ref19, ref20, ref21, ref22, ref23, ref24, ref25, ref26, ref27, ref28, ref29, ref30, ref31, ref32, refcy, zl3, mrc}. The unordered nature and rotational invariance of point clouds make it challenging to employ methods similar to traditional CNNs for feature extraction, and kNN is a commonly adopted approach.

\noindent{\textbf{Parametric point cloud analysis. }}The most extensive application of these methods lies in segmentation and primitive fitting \cite{hpnet, parsenet, spfn, h3d}. CAD shapes are typically composed of regular primitives, such as plane, cylinder, cone, and the variety of such elements is limited. Therefore, leveraging primitives' distinct shape features enables these methods to achieve superior segmentation and primitive fitting outcomes compared to general approaches. These methods generally predicting primitives parameters, and subsequently fed into downstream modules, these modules may incorporate specific primitive fitting algorithms. Despite their superior ability to exploit the shape features of individual primitives, these methods overlooked the constraints between primitives. This oversight limited their ability to distinguish CAD shapes that share similar appearances but serve different functions. Moreover, the primitive parameters predicting procedure necessitates the corresponding parameter labels. In practical, when sampling point clouds from CAD shapes, only coordinates can be obtained in most cases, which limits the applicability of these methods.

\noindent{\textbf{Other CAD-specific studies. }}
BRep-based methods typically involve BRep data conversion and feature extraction, the challenge lies in converting BRep into deep learning compatible format. These methods are commonly applied to classification \cite{ref33}, segmentation \cite{ref34, ref35}, assembly \cite{ref36, ref37, skh2jew}, engineering sketch generation \cite{ref38}, and CAD operation sequence prediction \cite{ref39, ref40}. 
Mesh-based methods use triangle models as input, such as STL and OBJ, contributions often rely on mechanical datasets provided by the authors. These methods are primarily used for classification \cite{ref41}, retrieval \cite{ref42}, segmentation \cite{ref43}, and BRep generation \cite{ref44}. 
Sketch-based learning takes free-hand sketches as input, with outputs typically being engineering sketches \cite{ref44, ref45, ref46, ref47, refzc}, 3D models \cite{ref48, ref49, zl1}, or CAD operation sequences \cite{ref50, sk2seq, dacad}. A key challenge is incorporating the regularity and connectivity of primitives, which is typically overlooked in graphic object analysis.

While there has been substantial progress in CAD-specific studies, further innovation on constraint learning is demanded to ensure higher accuracy and robustness.

\section{Method}

\subsection{Is Constraint Important?}
\label{sec:prob_state}

\begin{figure}[ht]
  \centering
  \includegraphics[width=0.8\linewidth]{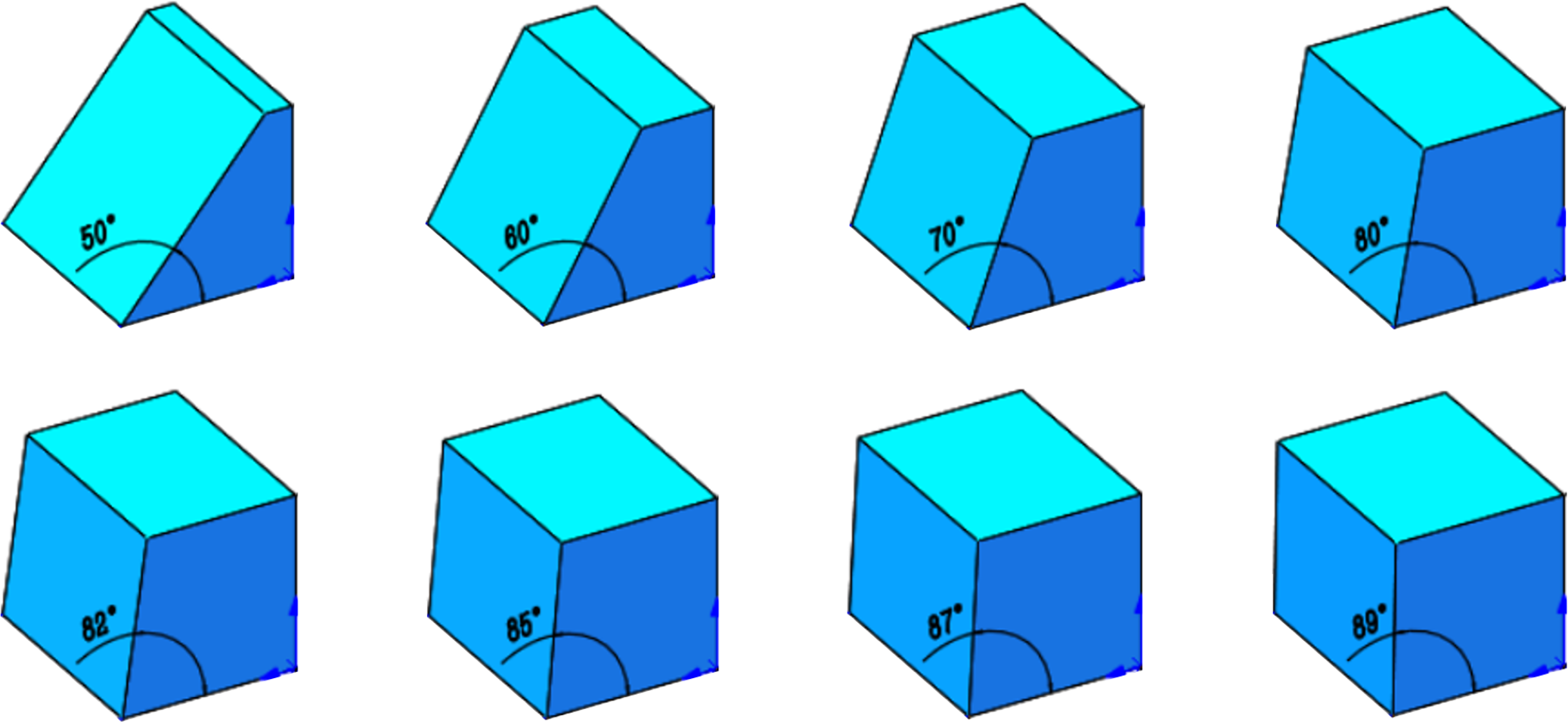}
  \caption{\textbf{Prisms compared to cuboid.}}
  \label{fig:prisms}
\end{figure}

\noindent To answer whether constraints could enhance deep learning methods' performance, we designed the following validation experiments. The objective is to discriminate prisms shown in \cref{fig:prisms} with cuboids. In each experiment, only prisms with a specific angle are selected, creating a binary classification task between prisms and cuboids. Consequently, eight independent experiments are conducted to cover all angles. As the prism’s angle approaches \ang{90}, its resemblance to cuboids increases, making it more challenging to distinguish them. The constraint-aware model is built on PointNet++ \cite{ref2} backbone, which predicts constraints and then adopts it for classification. The constraints are represented as the relation between the point attached primitive and a reference plane, more details in $Suppl.$

\begin{figure}[ht]
  \centering
  \includegraphics[width=1\linewidth]{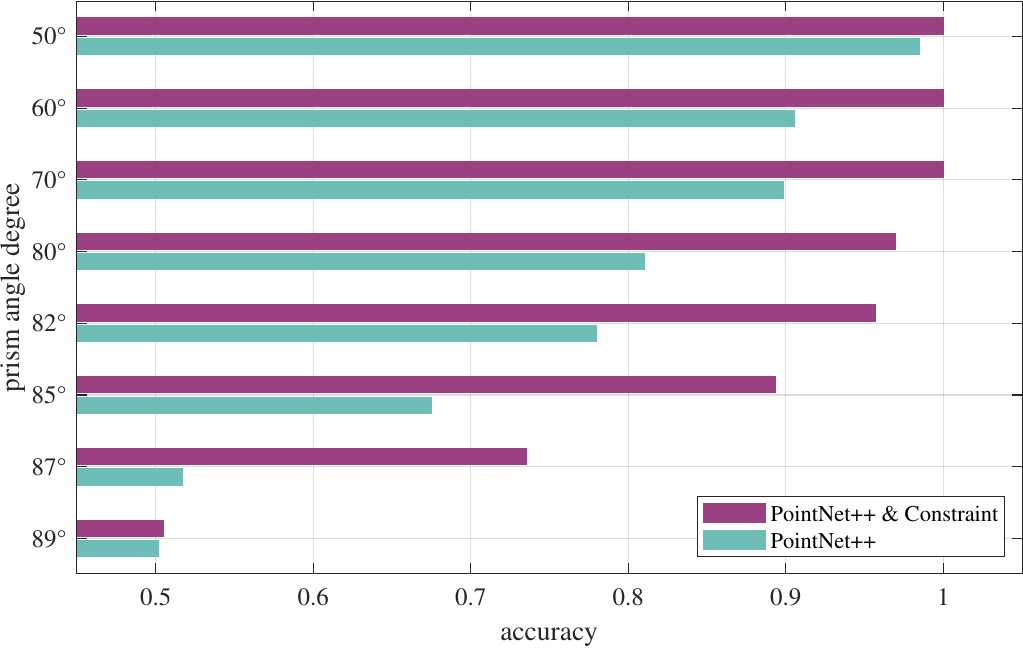}
  \caption{\textbf{Classification accuracy of validation experiments.}}
  \label{fig:validexp_res}
\end{figure}

From results presented in \cref{fig:validexp_res}, for prisms with angles from \ang{50} to \ang{87}, incorporating constraints improves classification accuracy significantly, while the \ang{89} prisms are too similar to the cuboids, all methods cannot distinguish them.

However, the constraint-aware learning method designed in the validation experiments is only applicable to simple scenarios. To develop a more generalized approach, we further investigated the constraint representation and deep learning architecture tailored for parametric point clouds.

\subsection{Constraint Representation}

\noindent In general, constraint representation should be defined as point-wise attribute for point clouds, analogous to normal vector. As point cloud is inherently unordered, assigning constraints at the point-wise level ensures that their representation is independent of point permutation. Following the constraint definitions in CAD software is not applicable. However, once the constraint is applied, the positional relations between primitives become fixed. Therefore, if the type of primitives and their relations are determined, the constraint representation on CAD shapes can be defined.

\begin{figure}[ht]
  \centering
  \includegraphics[width=0.9\linewidth]{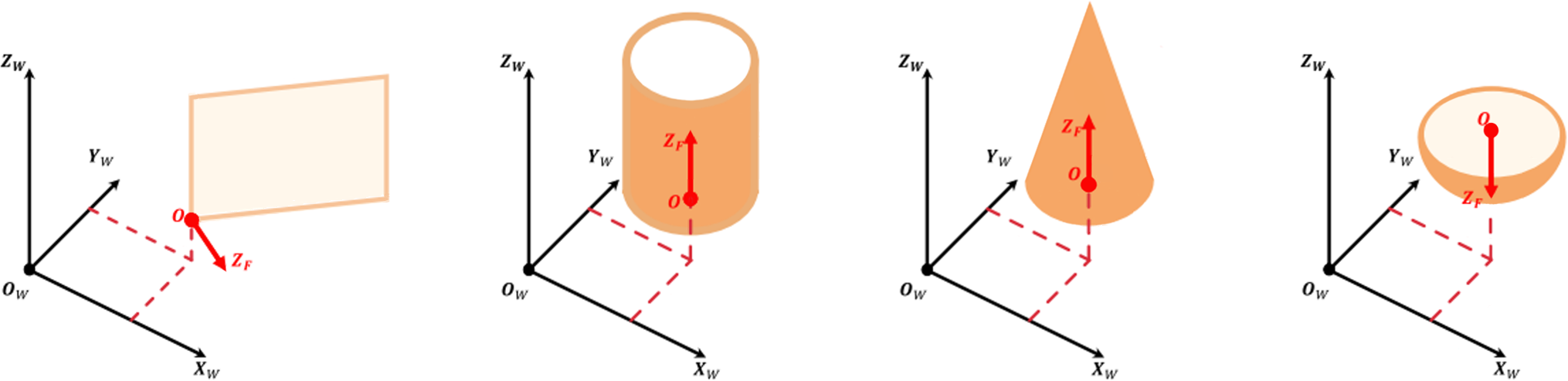}
  \caption{\textbf{MAD of primitives.} Plane: normal vector. Cylinder, cone, sphere: rotation axis. MAD computation requires unique process, details are provided in $Suppl.$}
  \label{fig:local_coor}
\end{figure}

\begin{figure}[ht]
  \centering
  \includegraphics[width=1.0\linewidth]{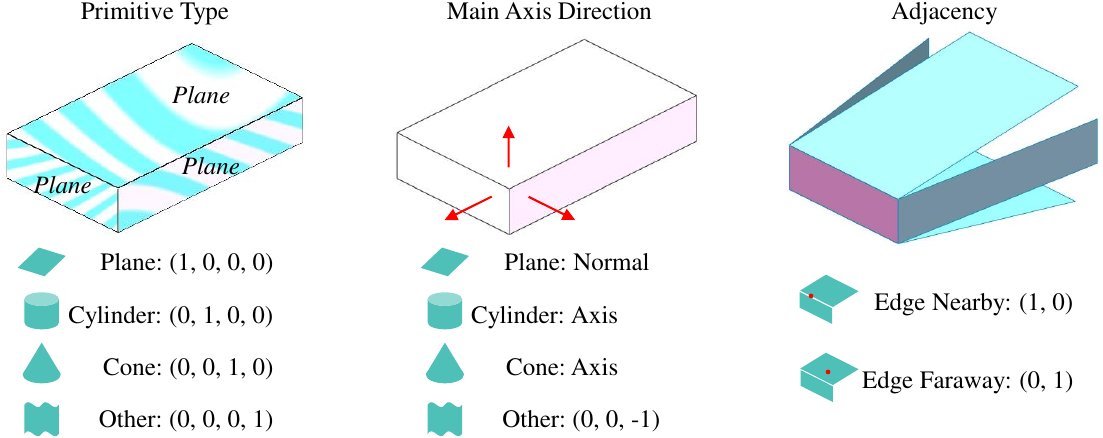}
  \caption{\textbf{Constraint representation of parametric point cloud.}}
  \label{fig:cst_rep}
\end{figure}

The point-wise primitive type can be represented as the type of the primitive to which the point is attached, using one-hot encoding. 
The positional relation is defined as the combination of main axis direction and adjacency. Considering that the direction of primitives' main axis is unique, as shown in \cref{fig:local_coor}, relations such as parallel and vertical could be derived from their main axes. However, some dimensional relations can not, such as the cylinders’ radius equivalence. 
Since the connected primitives are the basis of CAD shapes, the adjacency and main axis direction impose limitations on the distance and size of primitives. For example, in a cuboid, the adjacency and normal of the six faces ensure that the distance between the top and bottom faces equals the height of the front face, and the size of the back face is the same as the front one. Therefore, if the adjacency is combined with main axis directions, the positional relation can be determined comprehensively.
Adjacency represents the connection between primitives and cannot be expressed as point-wise attribute directly. The edges are key indicators of adjacency, with each edge corresponding to two adjacent primitives, adjacency can therefore be inferred from edges. For point cloud, an edge can be represented by marking points near it, thus, point-wise adjacency is expressed as whether the point is near an edge, which can be represented by one-hot vector. 
In summary, constraint representation in parametric point cloud is illustrated in \cref{fig:cst_rep}, and each point with constraint is expressed as $(x, y, z, MAD, Adj, PT)$.

Is MAD-Adj-PT sufficient to handle most constraints? For PT, commercial CAD systems typically classify primitives into plane, cylinder, cone, sphere, and free-form surface. Among these, plane, cylinder, and cone are the most common primitives, \eg, for a Trunk of ABC \cite{ref51}, plane: 69.63\%, cylinder: 17.51\%, cone: 6.99\%, other : 5.87\%. Besides that, constraints usually applied to regular primitives, while free-form surfaces rely heavily on geometric features. Therefore, considering plane, cylinder, and cone is sufficient for common constraint representation. For free-form surfaces, the CstNet Stage 2 also incorporates geometric features. MAD corresponds to directional constraints (\eg parallel, vertical), and MAD combined with PT can represent dimensional constraints (\eg distance, tangent). 
Through MAD and Adj, most common constraint types can be covered. Therefore, the combination of MAD, Adj, and PT is sufficient to represent the majority of constraints.

\subsection{Constraint Acquisition}
\begin{figure*}[ht]
  \centering
  \includegraphics[width=\linewidth]{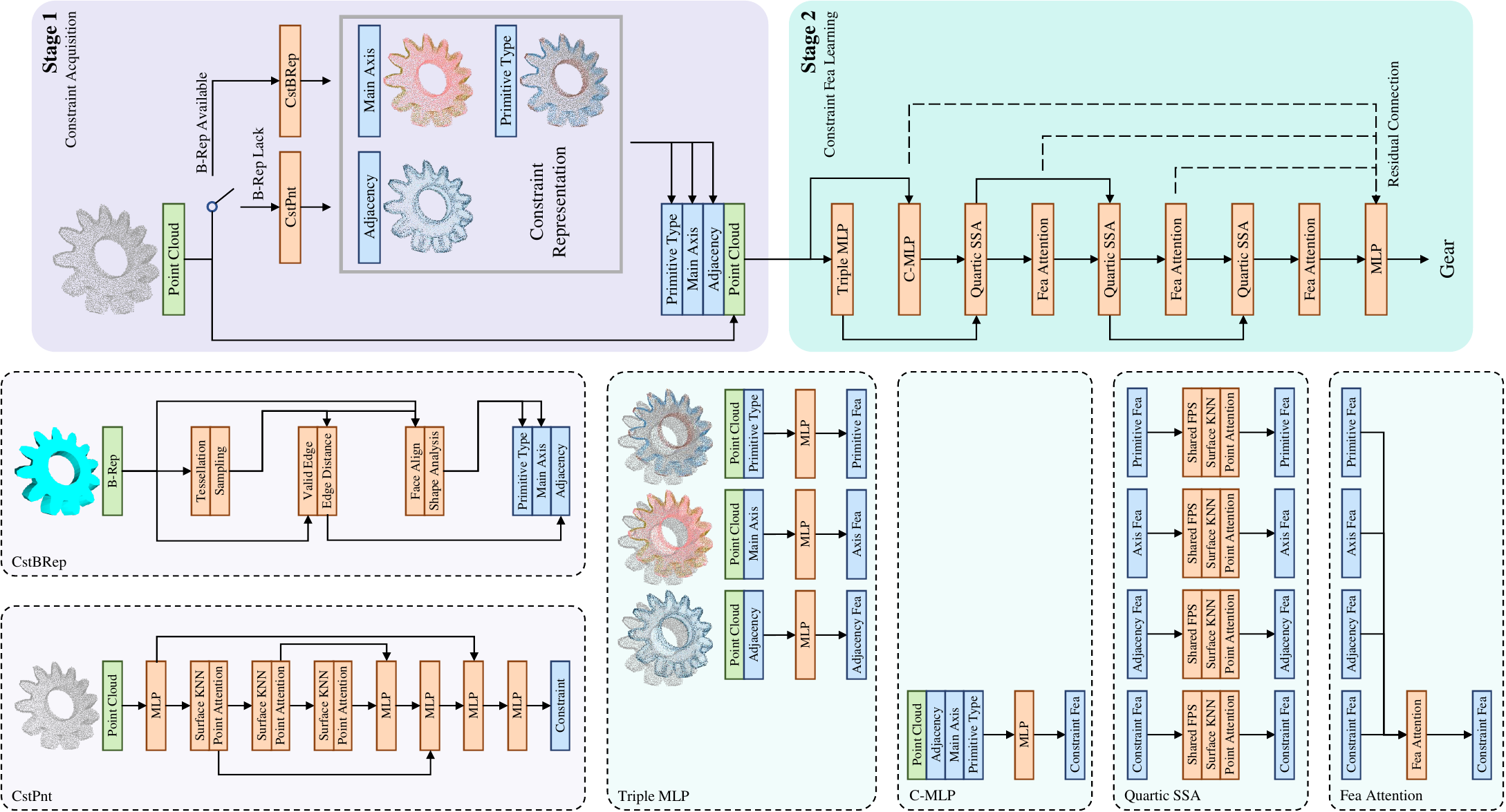}
  \caption{\textbf{Overview of CstNet.} The top side illustrates the overall architecture of CstNet, which comprising two stages. Stage 1 is designed for constraint acquisition. When BRep data is available, the CstBRep module is utilized to extract constraints; otherwise, use the CstPnt module. Stage 2 performs constraint feature learning, facilitating a deeper understanding of CAD shapes. The bottom side presents the details of module design. In the CstPnt module, the final MLP consists of three parallel MLPs with the same input, where the outputs correspond to Main Axis Direction, Adjacency, and Primitive Type.}
  \label{fig:solution_oview}
\end{figure*}

\noindent The Stage 1 of CstNet is designed to obtain constraint representation, as shown in the left of \cref{fig:solution_oview}. When BRep data is available, the constraint computation from BRep (CstBRep) module is selected. 
For the input BRep data, OCCT \cite{occt} performs tessellation to generate a mesh model, followed by Poisson disk sampling \cite{ref52} via pymeshlab to ensure a uniform point distribution and control the point count. If point cloud generation is not needed, this step could be skipped. 
For Adj, OCCT identify valid edges firstly. An edge is considered invalid if the two adjacent primitives are smoothly connected along the edge. Next, computing the distance from each point to all valid edges. A point is marked as near an edge if the minimum distance is below a threshold (we use $0.08 \times ShapeArea / 4 \pi$). For the PT and MAD, OCCT first determine which primitive it attached, and then analyze the primitive, providing both the PT and MAD.

In the absence of BRep data, such as working with mesh or voxel models, it can be converted to point cloud, after that the constraint prediction from point cloud (CstPnt) module can be used. Since each point's MAD-Adj-PT are local information, the computation only requires data from its neighboring points, meaning that global information is unnecessary. The MAD-Adj-PT computation of a point $\boldsymbol{p}$ could be divided into the following steps: 
\textbf{Step 1: }Identify neighbor points around $\boldsymbol{p}$. 
\textbf{Step 2: }From all neighbor points, identify which belongs to the same primitive as $\boldsymbol{p}$, store in $\boldsymbol{\mathrm{P_{valid}}}$. 
\textbf{Step 3: }Fit shapes such as cylinders or planes using $\boldsymbol{\mathrm{P_{valid}}}$, and determine the primitive type with the smallest fitting error. 
\textbf{Step 4: }Based on $\boldsymbol{\mathrm{P_{valid}}}$ and the primitive type, calculate the main axis direction and assess whether the point is near an edge. 
While the above process can be carried out by traditional methods, it is highly complex, therefore CstPnt is designed to achieve the same.

\begin{figure}[ht]
  \centering
  \includegraphics[width=0.95\linewidth]{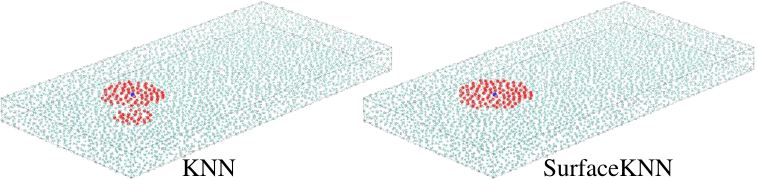}
  \caption{\textbf{Comparison of KNN and SurfaceKNN.} Blue point: center of neighbor search, red point: searched neighbors.}
  \label{fig:comp_knns}
\end{figure}

The design of CstPnt module is inspired by the above steps, with details in \cref{fig:solution_oview}. For \textbf{Step 1}, the k-Nearest Neighbors (KNN) algorithm is commonly used, which identifies neighbors within a sphere. However, for MAD-Adj-PT computation, it is more effective to search for neighbors along the shape’s surface. This approach captures more relevant points while reducing the influence of redundant points, as shown in \cref{fig:comp_knns}. To this end, we proposed a neighbor searching method SurfaceKNN, details in $Suppl.$
For \textbf{Step 2}, the attention mechanism between points is employed to approximate this process, which applies different weights to neighbor points when updating the center point’s embedding, as shown in \cref{eq:attn1}.
\textbf{Step 3} and \textbf{Step 4} can be accomplished by MLPs, we employed a U-Net-like structure for better data utilization.

\begin{equation}
  \boldsymbol{f}^{\prime}_i=
  \sum_{f_j\in\mathcal{N}_i}
  \rho
  \Bigl(
    \rm{MLP}
    \bigl(
    \mathrm{Q}(\boldsymbol{f}_{\mathit{i}})
    -\mathrm{K}(\boldsymbol{f}_{\mathit{j}})
    \bigr)
  \Bigr)
  \odot
  \mathrm{V}(\boldsymbol{f}_{\mathit{i}}).
  \label{eq:attn1}
\end{equation}

Where $\boldsymbol{f}_{\mathit{i}}$ is the feature of $\boldsymbol{p}_{i}$, $\mathcal{N}_i$ is the feature collection of all neighbor points around $\boldsymbol{p}_{i}$, $\rho$ is the normalization function (softmax in this paper), and $\odot$ denotes element-wise multiplication. $\mathrm{Q}(\boldsymbol{f})$, $\mathrm{K}(\boldsymbol{f})$, and $\mathrm{V}(\boldsymbol{f})$ are defined in similar structure by concatenating $\boldsymbol{f}$ with the positional encoding $\Delta$, and then feeding it into MLPs, where $\Delta$ is defined as $\Delta=\boldsymbol{p}_{i}-\boldsymbol{p}_{j}$.

The structure of CstPnt reveals that only local information is used, which aligns with the local nature of the constraint representation we designed. Since no global information is incorporated and most CAD shapes consist of a limited set of primitives, such as plane, cylinder, and cone, the CstPnt could generalize to unseen datasets after pre-training on a large dataset. The training data is derived from unlabeled BRep data, as a result, our CstPnt enables the effective use of existing large unlabeled BRep datasets, making it possible to serve meaningful tasks.

\subsection{Constraint-Aware Feature Learning}
\noindent The Stage 2 of CstNet is designed to leverage constraints effectively, which extracts features from MAD-Adj-PT separately, and attention between features is then applied to update constraint feature, with details in \cref{fig:solution_oview}.

The Triple MLP takes the coordinates, MAD, Adj, and PT as inputs. In this module, the coordinates are concatenated with three constraint components separately, and then feeds into three individual MLPs. The outputs are the Axis Feature, Adjacency Feature, and Primitive Feature, respectively. 
In the C-MLP layer, coordinates are concatenated with MAD-Adj-PT, and then feeds into MLPs, with the output being the initial Constraint Feature.
The Quartic SSA layer takes the Axis Feature, Adjacency Feature, Primitive Feature, and Constraint Feature as input. This layer consists of four paths, each processing the corresponding feature independently. In each path, Farthest Point Sampling (FPS) is performed to reduce the point scale. Subsequently, SurfaceKNN is used to search for neighbors of each sampled point. 
The outputs of FPS and SurfaceKNN are shared across the four paths. 
Finally, the point-level attention layers are employed to update the sample points' feature by its neighbors. The point-level attention here with a slight modification in the positional encoding $\Delta$ compare with \cref{eq:attn1}. For the path takes xyz \& MAD as input, $\Delta=[(\boldsymbol{p}_{i}-\boldsymbol{p}_{j} )\Vert(\boldsymbol{n}_{i}-\boldsymbol{n}_{j})]$, where $\boldsymbol{n}_{i}$, $\boldsymbol{n}_{j}$ represents the MAD of $\boldsymbol{p}_{i}$ and $\boldsymbol{p}_{j}$. For the paths that use xyz \& (Adj or PT) as input, $\Delta=[(\boldsymbol{p}_{i}-\boldsymbol{p}_{j})\Vert \boldsymbol{a}_{i} \Vert \boldsymbol{a}_{j}]$, where $\boldsymbol{a}_{i}, \boldsymbol{a}_{j}$ is the one-hot encoding of Adj or PT, $[\cdot \Vert \cdot]$ indicates concatenation in feature channel. 
Through Sampling, neighbor Searching, and Attention (SSA), the point cloud is down-scaled and features are updated.
The Fea Attention layer also takes the Axis Feature, Adjacency Feature, Primitive Feature, and Constraint Feature as input, feature-level attention is employed to update the constraint feature efficiently. This design considers that different points may focus on different features, for example, points near edge may focus more on the Adjacency Feature, while others may pay more attention to the Primitive or Axis Feature. The Fea Attention is developed by vector attention as formulated in \cref{eq:attn2}.

\begin{equation}
  \boldsymbol{f}^{\prime}_{\rm{cst}}=
  \sum_{\boldsymbol{f}_{j} \in \mathcal{F}}
  \rho
  \Bigl(
    \rm{MLP}
    \bigl(
    \mathrm{Q}(\boldsymbol{f}_{\rm{cst}})
    -\mathrm{K}(\boldsymbol{f}_{\mathit{j}})
    \bigr)
  \Bigr)
  \odot
  \mathrm{V}(\boldsymbol{f}_{\rm{cst}}).
  \label{eq:attn2}
\end{equation}

In which $\mathcal{F}=\{ \boldsymbol{f}_{\rm{mad}}, \boldsymbol{f}_{\rm{adj}}, \boldsymbol{f}_{\rm{pt}}, \boldsymbol{f}_{\rm{cst}} \}$, corresponding to Axis Feature, Adjacency Feature, Primitive Feature, and Constraint Feature. $\Delta$ is the one-hot encoding of $\boldsymbol{f}_{j}$ type.

For downstream classification tasks, the constraint features obtained in each module will be concatenated by residual connection and feed into maxpoling, MLPs and softmax. Constraint features obtained by different modules may differ in features' number and channel, based on the coordinates of the final constraint features, we identify the correspondence from different modules for further processing.

\section{Experiments}

\subsection{Experimental Settings}

\noindent We train CstNet using the Negative Log Likelihood Loss and optimized by Adam optimization for 200 epochs, with a weight decay of 0.0001 and an initial learning rate of 0.0001. We use StepLR learning rate scheduler, with step size of 20 and gamma of 0.7. The batch size is 16. All experiments are performed on GeForce RTX 4090 GPU.

\noindent \textbf{Param20K dataset: }
Param20K contains 19,739 multi-modal instances categorized into 75 classes. Each instance includes BRep data, mesh, and point cloud. All BRep data are collected from the following three sources. More details in the supplementary.
\textbf{Sources 1} (36.05\%): Downloaded and purified from TraceParts \cite{ref53}. The CAD shapes on TraceParts are collected from manufacturing companies.
\textbf{Sources 2} (24.12\%): Provided by a CAD company \cite{nd} that collaborates with our team. These shapes are all mechanical components.
\textbf{Sources 3} (39.83\%): Instantiated from parametric templates designed by our team. These templates are designed based on common mechanical structures. 

Our Param20K dataset encompasses various common formats of CAD shapes, making it suitable for a wide range of tasks. The careful selection and purification process undertaken by our team further ensures its high quality.

\subsection{Constraint Acquisition}
\begin{figure*}[htbp]
  \centering
  \includegraphics[width=1.0\linewidth]{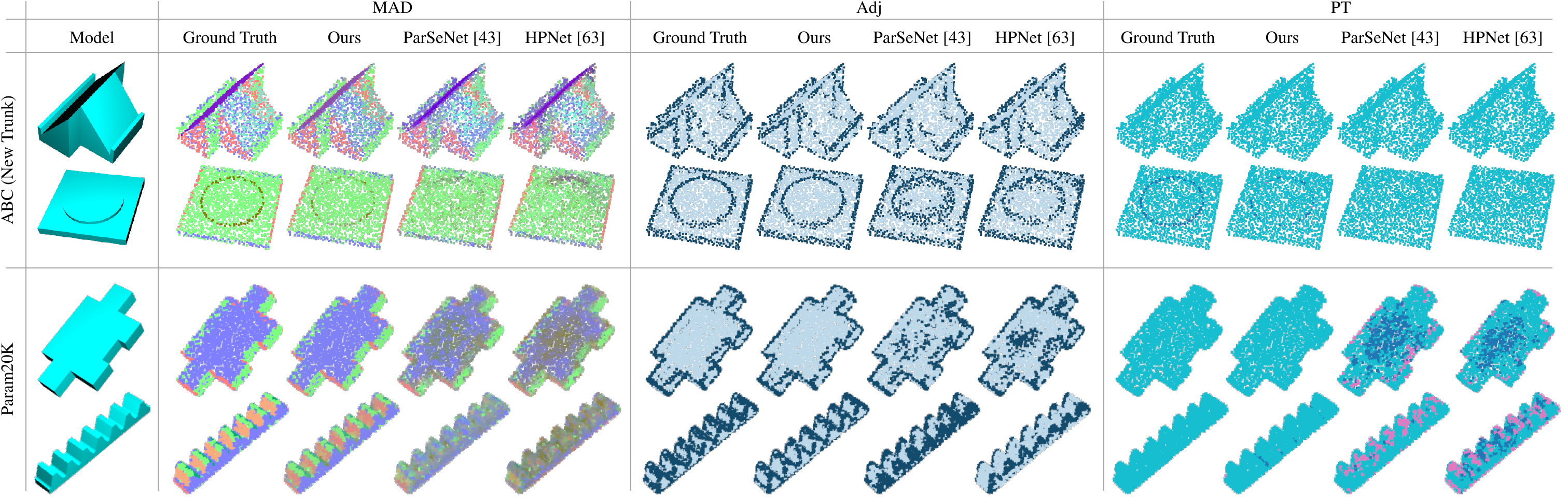}
  \caption{\textbf{Constraint prediction visualization on ABC and Param20K.} MAD is processed for visualization, details in $Suppl.$}
  \label{fig:abc_p20k}
\end{figure*}

\noindent Extracting MAD-Adj-PT from BRep data is accomplished by the CstBRep module, with results visualized in the Ground Truth column of \cref{fig:abc_p20k}. By referring to the model row, CstBRep performs satisfactorily in both MAD-Adj-PT extraction.
Predicting MAD-Adj-PT from point cloud is achieved by the CstPnt module. The training set is sourced from 25 trunks of ABC \cite{ref51}, each trunk containing 10,000 BRep files. We utilized the CstBRep to extract constraints as Ground Truth for model training. 
After the training finished, the MSE loss of MAD is 0.0559, the prediction accuracy of Adj and PT reached 94.62\% and 94.23\%, respectively.
In subsequent experiments, the above pre-trained CstPnt is utilized to predict MAD-Adj-PT with its weights frozen and no further updates. To ensure that only point cloud is used as input across all methods when comparing, even when BRep data is available, the CstBRep module will not be activated to obtain MAD-Adj-PT.

After CstPnt was pre-trained, we evaluated it on a ABC new trunk and Param20K dataset. The ABC new trunk was treated as the test set, while the Param20K was considered as an unseen CAD shapes dataset, more results on other public available datasets in $Suppl.$ Since there is no existing model that directly predicts MAD-Adj-PT, we modified the published CAD-specific methods \cite{hpnet, parsenet} by adding MLPs as prediction head to enable these predictions. The experiment results are shown in \cref{tab:cst_pred}, from which our CstPnt achieves the highest accuracy. Furthermore, from ABC to Param20K, all methods' performance decreased, but our method dropped least accuracy. ParSeNet and HPNet may be negatively affected by global features, which impaired their performance on unseen datasets. 
\cref{fig:abc_p20k} presents the visualization of constraint prediction results, which shown that our method is closest to the ground truth. Furthermore, from ABC to Paeam20K, our method demonstrates excellent generalization performance on unseen datasets. In contrast, while ParSeNet and HPNet exhibit acceptable prediction results on the ABC new trunk, they performed unsatisfactory on the unseen Param20K dataset.

\begin{table}[b]
\centering
\caption{\textbf{Constraint prediction on ABC (new trunk) and Param20K.} MAD: MSE loss of Main Axis Direction. Adj, PT: prediction accuracy of Adjacency and Primitive Type.}
\begin{tabularx}{\linewidth}{p{1.2cm} p{2.0cm}
                            >{\centering\arraybackslash}X
                            >{\centering\arraybackslash}X
                            >{\centering\arraybackslash}X}
\toprule
DataSet & Method & MAD$\downarrow$ & Adj$\uparrow$ & PT$\uparrow$ \\
\midrule
\multirow{3}*{ABC} & ParSeNet \cite{parsenet} & 0.1356 & 84.89 & 85.22 \\
~ & HPNet \cite{hpnet} & 0.1576 & 82.63 & 79.15 \\
~ & Ours & \textbf{0.0998} & \textbf{90.24} & \textbf{91.19} \\
\midrule
\multirow{3}*{Param20K} & ParSeNet \cite{parsenet} & 0.2247 & 81.42 & 59.75 \\
~ & HPNet \cite{hpnet} & 0.2570 & 78.60 & 57.66 \\
~ & Ours & \textbf{0.1390} & \textbf{87.95} & \textbf{86.52} \\
\bottomrule
\end{tabularx}
\label{tab:cst_pred}
\end{table}

\subsection{Classification}
\begin{figure}[htbp]
  \centering
  \includegraphics[width=1.0\linewidth]{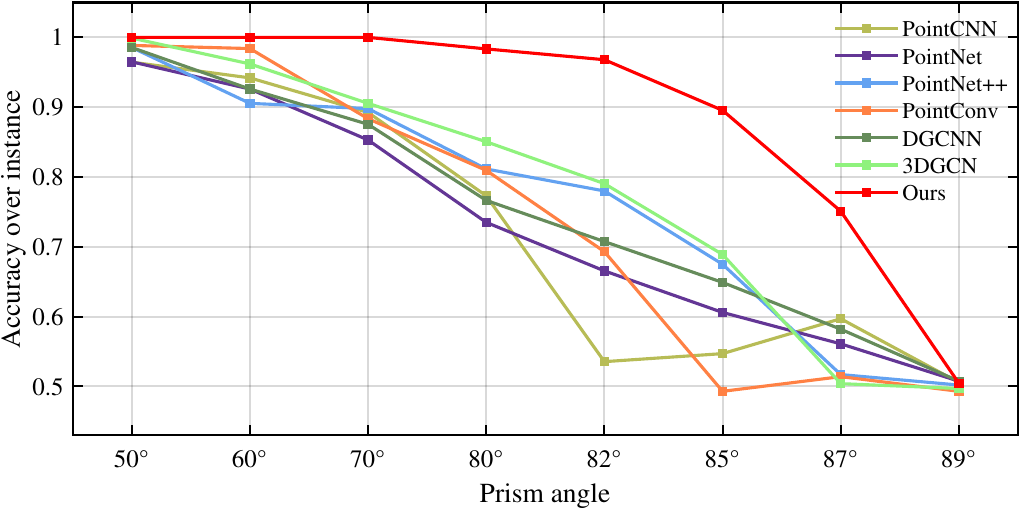}
  \caption{\textbf{Classification results on Prism-Cuboid dataset.}}
  \label{fig:cls_angle}
\end{figure}

\noindent To validate the ability of our CstNet in identifying CAD shapes that are visually similar but different in constraints, we conducted experiments on the Prism-Cuboid datasets presented in \cref{sec:prob_state}. Results shown in \cref{fig:cls_angle}, in which our CstNet achieved the best instance accuracy in distinguishing \ang{50} to \ang{87} prisms and cuboids, which demonstrates the strong ability of CstNet in discriminating CAD shapes with similar appearance but constraints different. For the binary classification of \ang{89} prisms and cuboids, all methods achieved approximately 50\% instance accuracy, which is attributed to the differences between them being so subtle that no one is capable of distinguishing them.

The classification of prisms and cuboids is predominantly influenced by constraints, while the comprehension of common CAD shapes necessitates the integration of both constraint and geometric features. CstNet achieves this by concatenating xyz with MAD-Adj-PT in Stage 2, thereby integrating geometric features into constraint features. To evaluate CstNet's performance on common CAD shapes, we conducted experiments on Param20K dataset, more results on other public available datasets in $Suppl.$ Results are presented in \cref{tab:cls_e20K}, where our CstNet outperformed other methods across both metrics, validated the effectiveness of utilizing constraint and geometric features.

\begin{table}
\centering
\caption{\textbf{Classification results on Param20K dataset.} Acc: accuracy over instance \%, Acc*: accuracy over class \%, F1: F1-score, mAP: mean average precision \%.}
\begin{tabularx}{1.0\linewidth}{p{2.3cm} >{\centering\arraybackslash}X >{\centering\arraybackslash}X >{\centering\arraybackslash}X >{\centering\arraybackslash}X}
\toprule
Method & Acc & Acc* & F1 & mAP \\
\midrule
PointCNN \cite{ref3}  & 80.25 & 76.03 & 74.64 & 83.38 \\
PointNet \cite{ref1}  & 81.30 & 83.21 & 82.06 & 85.18 \\
PointConv \cite{ref32}  & 82.30 & 85.62 & 84.91 & 87.92 \\
PointNet++ \cite{ref2}  & 83.70 & 86.37 & 85.30 & 87.94 \\
HPNet \cite{hpnet} & 83.85 & 80.87 & 80.95 & 87.80 \\
ParSeNet \cite{parsenet} & 83.96 & 79.50 & 78.52 & 85.66 \\
PTV2 \cite{ptv2} & 85.14 & 73.80 & 72.49 & 89.19 \\
DGCNN \cite{ref31}  & 85.40 & 87.28 & 86.43 & 89.17 \\
3DGCN \cite{ref30}  & 86.42 & 88.25 & 87.63 & 89.81 \\
PTMamba \cite{ptmamba} & 86.45 & 87.28 & 86.63 & 91.34 \\
Ours  & \textbf{89.94} & \textbf{91.06} & \textbf{90.34} & \textbf{92.72} \\
\bottomrule
\end{tabularx}
\label{tab:cls_e20K}
\end{table}

\begin{figure}[htbp]
  \centering
  \includegraphics[width=1.0\linewidth]{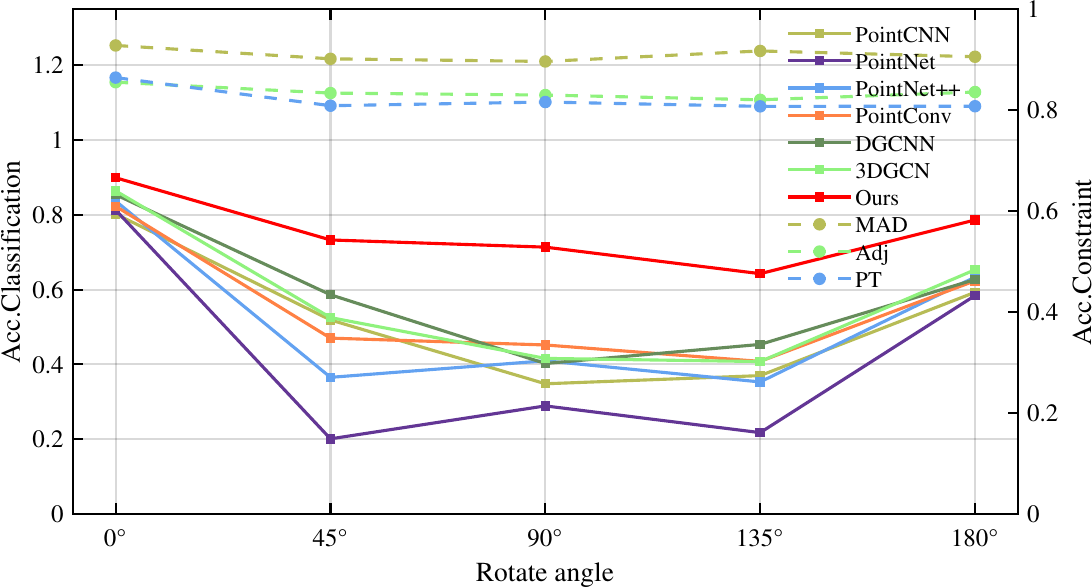}
  \caption{\textbf{Classification results on rotated Param20K.} The training set remained unchanged and the test set was rotated by a certain angle along the +Z direction. Solid line: left y axis, dashed line: right y axis. Acc.MAD = 1 - MAD.MSELoss.}
  \label{fig:cls_rot}
\end{figure}

To access the robustness of CstNet, we evaluated it under rotated point clouds. Since point clouds are pre-processed by aligning their centroids to the origin and scaling their ranges into a unit cube, the performance of all methods is unaffected by point cloud shifts and scaling. Comparison results are depicted in \cref{fig:cls_rot}, from which our CstNet demonstrates strongest robustness to point cloud rotation. The dashed line in \cref{fig:cls_rot} show that the performance of CstPnt module remains nearly unaffected by rotation, owing to it learned various local features from ABC dataset.


\subsection{Ablation Studies}

\noindent Our CstNet leverages all three components of constraint representation: MAD-Adj-PT. Additionally, we utilized the SurfaceKNN to obtain more efficient neighbors for each point. To validate the effectiveness of our design, we conducted a series of ablation studies on Param20K dataset.

For each component of constraint representation, we evaluated their importance by disabling one or all at a time. This operation is performed by setting the disabled components as a zero tensor. Classification results are presented in \cref{tab:ablation}, which show that all three components contribute to improving CstNet's performance. 
To assess the effectiveness of SurfaceKNN, we compared it with KNN. From \cref{tab:ablation}, it can be concluded that in most cases, our SurfaceKNN outperforms KNN. However, when the Primitive Type is disabled, the results are similar between the two methods, indicating that SurfaceKNN primarily enhances the effectiveness of the Primitive Feature.

\begin{table}
\centering
\caption{\textbf{Effects on varying neighbor search and constraint components.} Upper half: SurfaceKNN, lower half: KNN.}
\begin{tabularx}{\linewidth}{>{\centering\arraybackslash}X>{\centering\arraybackslash}X>{\centering\arraybackslash}X>{\centering\arraybackslash}X>{\centering\arraybackslash}X>{\centering\arraybackslash}X>{\centering\arraybackslash}X}
\toprule
\multicolumn{3}{c}{Constraint} & \raisebox{-0.5ex}{\multirow{2}*{Acc}} & \raisebox{-0.5ex}{\multirow{2}*{Acc*}} & \raisebox{-0.5ex}{\multirow{2}*{F1}} & \raisebox{-0.5ex}{\multirow{2}*{mAP}} \\
\cmidrule{1-3}
MAD & Adj & PT & & & & \\
\midrule
\ding{51} & \ding{51} & \ding{51} & \textbf{89.94} & \textbf{91.06} & \textbf{90.34} & \textbf{92.72} \\
\ding{55} & \ding{51} & \ding{51} & 86.32 & 89.05 & 88.38 & 90.59 \\
\ding{51} & \ding{55} & \ding{51} & 88.14 & 89.61 & 88.64 & 91.35 \\
\ding{51} & \ding{51} & \ding{55} & 88.68 & 88.94 & 89.01 & 91.22 \\
\ding{55} & \ding{55} & \ding{55} & 83.99 & 84.05 & 82.54 & 86.15 \\
\midrule
\ding{51} & \ding{51} & \ding{51} & \textbf{89.03} & \textbf{90.00} & \textbf{89.22} & \textbf{91.43} \\
\ding{55} & \ding{51} & \ding{51} & 85.04 & 85.93 & 85.74 & 88.08 \\
\ding{51} & \ding{55} & \ding{51} & 87.27 & 88.23 & 87.98 & 90.00 \\
\ding{51} & \ding{51} & \ding{55} & 88.73 & 88.97 & 88.91 & 91.21 \\
\ding{55} & \ding{55} & \ding{55} & 83.19 & 82.54 & 81.46 & 86.34 \\
\bottomrule
\end{tabularx}
\label{tab:ablation}
\end{table}


To validate the effectiveness of Stage 2 backbone, we conducted a comprehensive evaluation by feeding (x, y, z, MAD, Adj, PT) into multiple classification backbone for comparison. The results are presented in \cref{tab:abl_stg2}, from which CstNet demonstrates best performance under both predicted and label MAD-Adj-PT, indicating that our design achieves the best utilization of MAD-Adj-PT. Furthermore, although the accuracy of using predicted MAD-Adj-PT is lower than that of using label MAD-Adj-PT, it still shows a significant improvement compared to the baselines without MAD-Adj-PT. This highlights the effectiveness of the predicted MAD-Adj-PT in improving model performance.

\begin{table}
\centering
\caption{\textbf{Effectiveness of Stage2 Backbone.} Upper half: predicted constraint representation from CstPnt, lower: label constraint representation.}
\begin{tabularx}{1.0\linewidth}{p{2.3cm} >{\centering\arraybackslash}X >{\centering\arraybackslash}X >{\centering\arraybackslash}X >{\centering\arraybackslash}X}
\toprule
Backbone-S2 & Acc & Acc* & F1 & mAP \\
\midrule
PointNet \cite{ref1}  & 83.83 & 83.64 & 83.20 & 87.27 \\
PointNet++ \cite{ref2}  & 85.52 & 88.11 & 87.18 & 90.79 \\
DGCNN \cite{ref31}  & 86.81 & 88.70 & 87.70 & 91.35 \\
3DGCN \cite{ref30}  & 87.46 & 89.59 & 87.97 & 91.92 \\
Ours & \textbf{89.94} & \textbf{91.06} & \textbf{90.34} & \textbf{92.72} \\
\midrule
PointNet \cite{ref1}  & 86.24 & 85.99 & 86.33 & 91.00 \\
PointNet++ \cite{ref2}  & 86.60 & 88.82 & 87.71 & 90.74 \\
DGCNN \cite{ref31}  & 87.72 & 89.14 & 88.78 & 91.16 \\
3DGCN \cite{ref30}  & 89.29 & 89.81 & 89.00 & 92.62 \\
Ours  & \textbf{91.80} & \textbf{92.52} & \textbf{91.76} & \textbf{94.20} \\
\bottomrule
\end{tabularx}
\label{tab:abl_stg2}
\end{table}

\section{Conclusions}

\noindent In this paper, we studied the constraints of CAD shapes and introduced its deep learning-friendly representation. Afterward, the CstNet is proposed for extracting and leveraging constraint representation. Finally, we built a multi-modal CAD dataset Param20K. Extensive experiments demonstrate that the CstNet achieves SOTA performance on both accuracy and robustness. Comprehensive ablation studies further validate the effectiveness of the proposed constraint representation, SurfaceKNN, and network backbone.

CAD shapes possess numerous attributes. We analyze the constraints from a functional perspective in this work. In future work, we plan to investigate additional aspects such as part design and manufacturing procedure, to achieve a deeper understanding of CAD shapes.

\section*{Acknowledgements}
\noindent This work was supported by National Key Research and Development Program of China (Grant No.2022YFB3303101).

{
    \small
    \bibliographystyle{ieeenat_fullname}
    \bibliography{main}
}



\end{document}